%% file: eccv2022submission.tex
\documentclass[runningheads]{llncs}
\usepackage{graphicx}

\usepackage{tikz}
\usepackage{comment}
\usepackage{amsmath,amssymb} 
\usepackage{color}

\usepackage{algorithm}
\usepackage{algorithmic}
\usepackage{colortbl}

\usepackage{mathtools}

\usepackage{xcolor}

\definecolor{mygray}{gray}{.95}
\definecolor{mygreen}{RGB}{239,255,246}
\definecolor{mypurple}{RGB}{246,240,250}
\definecolor{myred}{RGB}{255,239,239}

\usepackage{multirow}
\newcommand{\tabincell}[2]{\begin{tabular}{@{}#1@{}}#2\end{tabular}}
\usepackage{makecell}

\usepackage{booktabs}

\usepackage{bbding}

\usepackage{pifont}

\usepackage{overpic}
\usepackage{bm}

\newcommand{\etal}{{\emph{et al.}}}

\usepackage{caption}

\usepackage{wrapfig}

\usepackage{hyperref}

\usepackage[accsupp]{axessibility}  

\begin{document}
\pagestyle{headings}
\mainmatter
\def\ECCVSubNumber{5410}  

\title{Switchable Online Knowledge Distillation} 

\titlerunning{Switchable Online Knowledge Distillation}
%
\author{Biao Qian\inst{1} \and
Yang Wang\inst{1,*} \and
Hongzhi Yin\inst{2} \and
Richang Hong\inst{1} \and
Meng Wang\inst{1}
}
\authorrunning{B. Qian et al.}
%
\institute{Key Laboratory of Knowledge Engineering with Big Data, Ministry of Education, School of Computer Science and Information Engineering, Hefei University of Technology, China
 \and
The University of Queensland\\
\email{\{hfutqian,hongrc.hfut,eric.mengwang\}@gmail.com \  yangwang@hfut.edu.cn \  h.yin1@uq.edu.au}
}
\maketitle

\begin{abstract}
Online Knowledge Distillation (OKD) improves the involved models by reciprocally exploiting the difference between teacher and student. Several crucial bottlenecks over the gap between them --- e.g., Why and when does a large gap harm the performance, especially for student? How to quantify the gap between teacher and student? --- have received limited formal study. In this paper, we propose \textbf{Swit}chable \textbf{O}nline \textbf{K}nowledge \textbf{D}istillation (SwitOKD), to answer these questions. Instead of focusing on the accuracy gap at test phase by the existing arts, the core idea of SwitOKD is to adaptively calibrate the gap at training phase, namely distillation gap, via a switching strategy between two modes --- expert mode (pause the teacher while keep the student learning) and learning mode (restart the teacher). To possess an appropriate distillation gap, we further devise an adaptive switching threshold, which provides a formal criterion as to when to switch to learning mode or expert mode, and thus improves the student's performance. Meanwhile, the teacher benefits from our adaptive switching threshold and keeps basically on a par with other online arts. We further extend SwitOKD to multiple networks with two basis topologies. Finally, extensive experiments and analysis validate the merits of SwitOKD for classification over the state-of-the-arts. Our code is available at \url{https://github.com/hfutqian/SwitOKD}.

\end{abstract}

\input{Introduction}

\input{technique}

\input{experiment}

\section{Conclusion}
In this paper,  we propose Switchable Online Knowledge Distillation (SwitOKD), to mitigate the adversarial impact of large distillation gap between teacher and student, where our basic idea is to calibrate the distillation gap by adaptively pausing the teacher to wait for the learning of student. We foster it throughout an adaptive switching strategy between learning mode and expert mode. Notably, an adaptive switching threshold is devised to endow SwitOKD with the capacity to automatically yield an appropriate distillation gap, so that the performance of student and teacher can be improved. Further, we verify SwitOKD’s extendibility to multiple networks. The extensive experiments on typical classification datasets validate the effectiveness of SwitOKD.

\noindent \textbf{Acknowledgements.} This work is supported by the National Natural Science Foundation of China under grant no U21A20470, 62172136, 61725203, U1936217. Key Research and Technology Development Projects of Anhui Province (no. 202004a5020043).

\clearpage
%
%
\bibliographystyle{splncs04}
\bibliography{egbib}
\end{document}

%% file: Introduction.tex
\section{Introduction}
\label{introduction}
The essential purpose of Knowledge Distillation (KD) \cite{hinton2015distilling,tung2019similarity,li2020few,passalis2020heterogeneous,yim2017gift,wang2018kdgan,menon2021statistical,zhu2021student,wang2021survey} is to improve the performance of a \emph{low-capacity student network} (small size, compact) for model compression by distilling the knowledge from a high-capacity teacher network (large size, over parameterized)$^1$\footnotetext{$*$ Yang Wang is the corresponding author. \\ $^1$ Throughout the rest of the paper, we regard high-capacity network as teacher and low-capacity network as student for simplicity.}.
The conventional knowledge distillation \cite{hinton2015distilling,xu2020knowledge,mirzadeh2020improved,Cho_2019_ICCV,jin2019knowledge,chen2021distilling,huang2021revisiting,zhu2021complementary} requires a pre-trained teacher to serve as the \emph{expert} network in advance, to be able to provide better supervision for the student in place of one-hot labels. 
However, it is usually a two-stage offline process, which is inflexible and requires extra computational cost.

\begin{figure}[t]
\centering
\setlength{\abovecaptionskip}{0.1cm}
\setlength{\belowcaptionskip}{-0.6cm}
\includegraphics[width=0.9\textwidth]{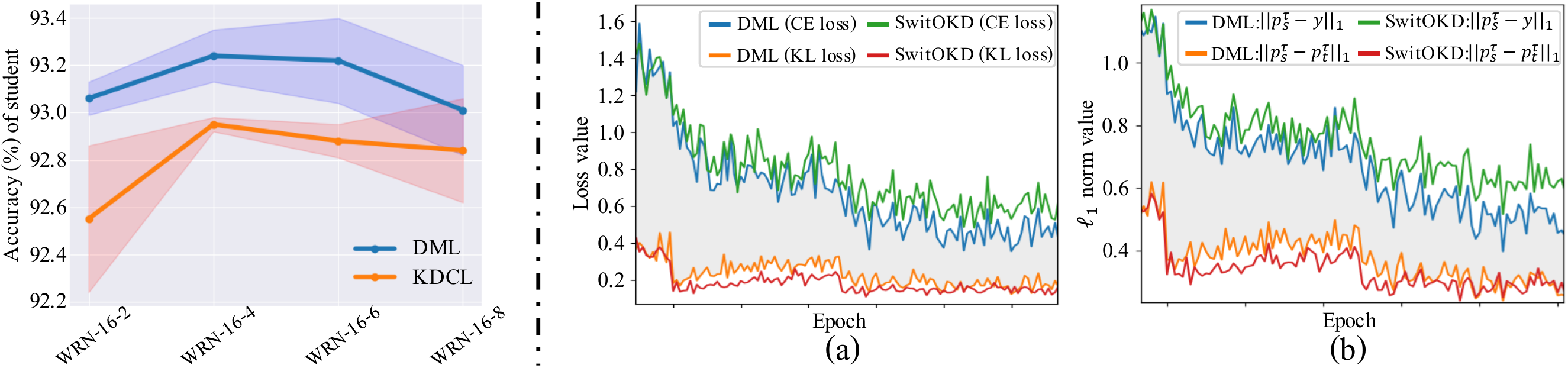}
\caption{ \textbf{Left:} Illustration of how the large accuracy gap between teacher (WRN-16-2 to WRN-16-8) and student (ResNet-20) affects online distillation process on CIFAR-100 \cite{krizhevsky2009learning}. \textbf{Right:} DML \cite{zhang2018deep} bears the emergency of escaping online KD under (a) large \emph{accuracy gap} and (b) large \emph{distillation gap}, whereas SwitOKD extends online KD's lifetime to avoid the degeneration. }
\label{dml_kdcl_gap}
\end{figure}

Unlike offline fashion, the goal of recently popular online knowledge distillation is to reciprocally train teacher and student from scratch, where they learn extra knowledge from each other, and thus improve themselves simultaneously \cite{zhang2018deep,chen2020online,chung2020feature,song2018collaborative}. 
Typically, Deep Mutual Learning (DML) \cite{zhang2018deep} encourages each network to mutually learn from each other by mimicking their predictions via Kullback Leibler (KL) divergence. 
Chen \etal \cite{chen2020online} presents to improve the effectiveness of online distillation by assigning weights to each network with the same architecture. Further, Chung \etal \cite{chung2020feature} proposes to exchange the knowledge of feature map distribution among the networks via an adversarial means.
Most of these approaches tend to \emph{equally} train the same or different networks with \emph{small accuracy gap}, where they usually lack richer knowledge from a powerful teacher.
In other words, online fashion still fails to resolve the problem of student's performance impairment caused by a large accuracy gap \cite{mirzadeh2020improved,Cho_2019_ICCV,jin2019knowledge} (see Fig.\ref{dml_kdcl_gap} Left), thus somehow violating the principle of KD.
As inspired, we revisit such long-standing issue, and find the fundamental \emph{bottlenecks} below:
1) when and how the gap has negative effect on online distillation process. For example, as the teacher turns from WRN-16-4 to WRN-16-8 (larger gap), the student accuracy rapidly declines (see Fig.\ref{dml_kdcl_gap} Left), while KL loss for the student degenerates into Cross-Entropy (CE) loss (see Fig.\ref{dml_kdcl_gap} Right(a)) as per loss functions in Table \ref{difference_recent}. To mitigate that, we raise 2) how to quantify the gap and automatically adapt to various accuracy gap, particularly large accuracy gap.

One attempt derives from Guo \etal \cite{Guo_2020_CVPR}, who studied the effect of large accuracy gap on distillation process and found that a large accuracy gap constitutes a certain harm to the performance of teacher. To this end, they propose KDCL, to admit the accuracy improvement of teacher by generating a high-quality soft target, so as to benefit the student.
Unfortunately, KDCL pays more attention to teacher, which deviates from the essential purpose of KD; see Table \ref{difference_recent}.

To sum up, the above online fashions overlook the principle of KD.
Meanwhile, they focus on the accuracy gap that is merely obtained at \emph{test} phase, which is not competent for quantifying the gap since it offers no guidance for the distribution alignment of distillation process performed at \emph{training} phase. For instance, the accuracy just depends on the class with maximum probability given a 10-class output, while distillation usually takes into account all of the 10 classes.
As opposed to them, we study the gap (the difference in class distribution between teacher and student) at \emph{training} phase, namely \emph{distillation gap}, which is quantified by $\ell_1$ norm of the gradient (see Sec.\ref{quantify_gap}), and how it affects online distillation process from student's perspective.
Taking DML~\cite{zhang2018deep} as an example, we observe that the gradient for KL loss $||p_s^{\tau} - p_t^{\tau}||_1$ increasingly degenerates into that for CE loss $||p_s^{\tau}-y||_1$ given a large gap; see Fig.\ref{dml_kdcl_gap} Right(b). In such case, the student suffers from \emph{the emergency of escaping online KD process}.

\begin{table*}[t]
\centering
\setlength{\belowcaptionskip}{0.1cm}
\caption{The varied loss functions for typical distillation methods and the common form of their gradients. $\tau$ is set to 1 for theoretical and experimental analysis. $ensemble$ is used to generate a soft target by combining the outputs of teacher and student. The gradient for KL divergence loss exactly reflects the difference between the predictions of student and teacher. 
}
\tiny
\begin{tabular}{c|c|c|c}
\toprule

\makecell[c]{Method}      & Loss function of the networks  &\tabincell{c}{The common form \\ of the gradient }     & \tabincell{c}{Focus on \\  \ student or not \  }                    \\
\hline\hline

\tabincell{c}{KD \cite{hinton2015distilling}  \\ \ (NeurIPS 2015) \ }           &\cellcolor{mygreen} \quad$\mathcal{L}=\alpha \mathcal{L}_{CE}(y,p_s^1) + (1-\alpha) \tau^2 \mathcal{L}_{KL}(p_t^{\tau},p_s^{\tau}) \quad$
& \cellcolor{mypurple} $(p_s^1-y)+(p_s^{\tau}-p_t^{\tau})$  &\cellcolor{myred}\ding{51}      \\

\hline\hline

\tabincell{c}{KDCL \cite{Guo_2020_CVPR}    \\ (CVPR 2020)}            &\cellcolor{mygreen} \tabincell{c}{ $\mathcal{L}=\sum\limits_{i} \mathcal{L}^i_{CE}(y,p_i^1) +  \tau^2 \mathcal{L}^i_{KL}(p_m,p_i^{\tau}), $ \\   $p_m=ensemble(p_s^{\tau},p_t^{\tau})$}
& \cellcolor{mypurple} \tabincell{c}{  $(p_i^1-y)+(p_i^{\tau}-p_m)$, \\ $i=s,t$ }   &\cellcolor{myred}\ding{55}     \\
\hline

\tabincell{c}{DML \cite{zhang2018deep}    \\ (CVPR 2018)}             &\cellcolor{mygreen}  \tabincell{c}{ $\mathcal{L}_s=\mathcal{L}_{CE}(y,p_s^1) +  \mathcal{L}_{KL}(p_t^{\tau},p_s^{\tau}), $ \\   $ \mathcal{L}_t=\mathcal{L}_{CE}(y,p_t^1) +  \mathcal{L}_{KL}(p_s^{\tau},p_t^{\tau})$}
& \cellcolor{mypurple} \tabincell{c}{ $(p_s^1-y)+(p_s^{\tau}-p_t^{\tau})$, \\ $(p_t^1-y)+(p_t^{\tau}-p_s^{\tau})$}  &\cellcolor{myred}\ding{55}            \\
\hline
\hline

\makecell[c]{\textbf{SwitOKD} \\ \textbf{(Ours)}}             &\cellcolor{mygreen}  \tabincell{c}{ $\mathcal{L}_s=\mathcal{L}_{CE}(y,p_s^1) + \alpha \tau^2 \mathcal{L}_{KL}(p_t^{\tau},p_s^{\tau}), $ \\  $p_t^{\tau}=p_t^{\tau,l}  \textcolor{red}{\Leftrightarrow}  p_t^{\tau}=p_t^{\tau,e}$ }
& \cellcolor{mypurple} \tabincell{c}{ \ \ $(p_s^1-y)+(p_s^{\tau}-p^{\tau,l}_t) \ \ $ \\ $\textcolor{red}{\Updownarrow}$  \\  $(p_s^1-y)+(p_s^{\tau}-p^{\tau,e}_t)$ }    &\cellcolor{myred}\ding{51}      \\

\bottomrule
\end{tabular}
\label{difference_recent}
\end{table*}

\begin{figure*}[t]
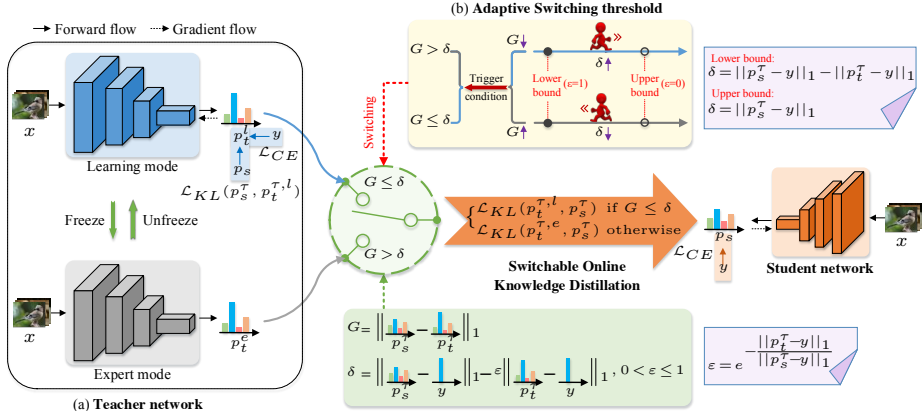

\centering
\setlength{\abovecaptionskip}{0.6cm}
\setlength{\belowcaptionskip}{0.1cm}

\begin{overpic}[width=1.0\textwidth]{overview2.pdf}

\put(2,30){\scriptsize{$x$}}
\put(2,7.5){\scriptsize{$x$}}
\put(97,18){\scriptsize{$x$}}

\put(27.5,27.8){\tiny{$\mathcal{L}_{CE}$}}
\put(29,29.8){\tiny{$y$}}
\put(24.8,29.6){\tiny{$p_t^{l}$}}
\put(24.8,26){\tiny{$p_s$}}
\put(19,23.5){\tiny{$\mathcal{L}_{KL}(p_s^{\tau},p_t^{\tau,l})$}}
\put(24.5,7){\tiny{$p_t^{e}$}}

\put(50,20.6){\scriptsize{$
\begin{cases}
  \\
  
\end{cases}$}}
\put(50.4,21.7){ \tiny{$\mathcal{L}_{KL}(p_t^{\tau,l},p_s^{\tau})$} }          \put(64.6,21.7){ \tiny{if $G\le \delta$} }
\put(50.4,19.3){ \tiny{$\mathcal{L}_{KL}(p_t^{\tau,e},p_s^{\tau})$} }         \put(64.6,19.3){ \tiny{otherwise} }

\put(76.5,36.5){\tiny{$\delta\!=\!||p_s^{\tau}\!-\!y||_1\!-\!||p_t^{\tau}\!-\!y||_1$}}
\put(76.5,32.5){\tiny{$\delta\!=\!||p_s^{\tau}\!-\!y||_1$}}
\put(76.5,4.5){\tiny{$\varepsilon\!=\!e^{-\!\frac{||p_t^{\tau}\!-\!y||_1}{||p_s^{\tau}\!-\!y||_1}}$}}

\put(77.3,18.8){\tiny{$p_s$}}
\put(77.7,15.3){\tiny{$y$}}
\put(72.7,17){\tiny{$\mathcal{L}_{CE}$}}

\put(44,39){\tiny{$G\!>\!\delta$}}  \put(54.5,40){\tiny{$G$}}  \put(64.2,37.5){\tiny{$\delta$}}
\put(44,31.3){\tiny{$G\!\le\!\delta$}}  \put(54.5,30){\tiny{$G$}}  \put(64.2,30){\tiny{$\delta$}}
\put(39,24.5){\tiny{$G \!\le\! \delta$}}
\put(39,16.6){\tiny{$G\!>\!\delta$}}

\put(37.2,8.7){\tiny{$G$}}
\put(37.2,3.8){\tiny{$\delta$}}
\put(42,7){\tiny{$p_s^{\tau}$}}  \put(47,7){\tiny{$p_t^{\tau}$}}  \put(50.8,8){\tiny{$1$}}  \put(50.8,3.2){\tiny{$1$}}  \put(64.8,3.2){\tiny{$1$}}
\put(42,1.8){\tiny{$p_s^{\tau}$}}  \put(47,1.8){\tiny{$y$}}  \put(53.2,4){\tiny{$\varepsilon$}}  \put(56,1.8){\tiny{$p_t^{\tau}$}}  \put(61,1.8){\tiny{$y$}}  \put(66,4){\tiny{$,0\!<\!\varepsilon \!\le\! 1$}}

\end{overpic}

\caption{Illustration of the proposed SwitOKD framework. Our basic idea is to adaptively pause the training of teacher while make the student continuously learn from teacher, to mitigate the adversarial impact of large distillation gap on student. Our framework is achieved by an adaptive switching strategy between two training modes: \emph{learning mode} that is equivalent to training two networks reciprocally and \emph{expert mode} that freezes teacher's parameters while keeps the student learning. Notably, we devise an adaptive switching threshold (b) to admit automatic switching between learning mode and expert mode for an appropriate distillation gap (quantified by $G$). See Sec.\ref{optim} for the detailed switching process.
}
\label{overview}
\end{figure*}

In this paper, we study online knowledge distillation and come up with a novel  framework, namely \textbf{Swit}chable \textbf{O}nline \textbf{K}nowledge \textbf{D}istillation(SwitOKD), as illustrated in Fig.\ref{overview}, which stands out new ways to mitigate the adversarial impact of large distillation gap on student.
The basic idea of SwitOKD is to calibrate the distillation gap by adaptively pausing the teacher to wait for the learning of student during the \emph{training} phase.
Technically, we specify it via an adaptive switching strategy between two types of training modes: namely \emph{learning mode} that is equivalent to reciprocal training from scratch and \emph{expert mode} that freezes teacher's weights while keeps the student learning. Notably, we devise an adaptive switching threshold to endow SwitOKD with the capacity to  yield an appropriate distillation gap that is conducive for knowledge transfer from teacher to student.
Concurrently, it is nontrivial to devise an ``ideal" switching threshold (see Sec.\ref{threshold_bound}) due to: 1) not too large --- a large threshold aggressively pushes \emph{learning mode} and enlarges the distillation gap, resulting the student into the emergency of escaping online KD process; such fact, as expanded in Sec.\ref{threshold_bound}, will further trap teacher to be paused constantly; as opposed to 2) not too small --- the teacher constantly performs \emph{expert mode} and receives poor accuracy improvement, suffering from no effective knowledge distilled from teacher to student. The above two conditions lead to 3) adaptiveness --- the threshold is adaptively calibrated to balance learning mode and expert mode for extending online KD’s lifetime.
\emph{Following} SwitOKD, we further establish two fundamental basis topologies to admit the extension of multi-network setting.
The extensive experiments on typical datasets demonstrate the superiority of SwitOKD.

%% file: technique.tex
\section{Switchable Online Knowledge Distillation}
\label{skd_method}

Central to our method are three aspects:
(i) quantifying the distillation gap between teacher and student, and analyzing it for online distillation (Sec.\ref{quantify_gap} and \ref{tradition});
(ii) an adaptive switching threshold to mitigate the adversarial impact of large distillation gap from student's perspective (Sec.\ref{threshold_bound}); and,
(iii) extending SwitOKD to multiple networks (Sec.\ref{multi_n}).

\subsection{How to Quantify the Distillation Gap Between Teacher and Student?}
\label{quantify_gap}
Thanks to varied random starts and differences in network structure (\emph{e.g.}, layer, channel, etc), the prediction difference between teacher and student always exists, which is actually exploited to benefit online distillation.
Since the accuracy obtained at \emph{test} phase is not competent to quantify the gap,
we propose to quantify the gap at \emph{training} phase, namely \emph{distillation gap}, by computing $\ell_1$ norm of the gradient for KL divergence loss (see Table \ref{difference_recent}), denoted as $G$, which is more suitable for capturing the same elements (0 entries) and the element-wise difference between the predictions of student and teacher, owing to the \emph{sparsity} property of $\ell_1$ norm.
Concretely, given a sample $x$, let $p_t^{\tau}$ and $p_s^{\tau}$ represent the softened outputs of a teacher network $\mathcal{T}(x,\theta_t)$ and a student network $\mathcal{S}(x,\theta_s)$, respectively, then $G$ is formulated as 
\begin{footnotesize} 
\begin{equation}
\setlength{\abovedisplayskip}{1pt}
\setlength{\belowdisplayskip}{1pt}
\label{gap_quantify}
\begin{aligned}
G&=||p_s^{\tau}-p_t^{\tau}||_1=\frac{1}{K}\sum^{K}_{k=1} |p_s^{\tau}(k)-p_t^{\tau}(k)|,G \in [0,2],
\end{aligned}
\end{equation}
\end{footnotesize}
where $|.|$ denotes the absolute value and $\tau$ is the temperature parameter. The $k$-th element of the softened output $p_f^{\tau}$ is denoted as $p_f^{\tau}(k)=\frac{exp(z_f(k) / \tau)}{\sum_j^K exp(z_f(j) / \tau)}, f=s,t$; $z_f(k)$ is the $k$-th value of the logit vector $z_f$. $K$ is the number of classes. 
Prior work observes that a great prediction difference between teacher and student has a negative effect on distillation process \cite{Cho_2019_ICCV,jin2019knowledge,Guo_2020_CVPR}.  
Next, we discuss how the distillation gap affects online distillation process from student's perspective.

\subsection{Why is an Appropriate Distillation Gap Crucial?}
\label{tradition}

It is well-accepted that knowledge distillation loss for student is the KL divergence of the soften outputs of teacher $p_t^{\tau}$ and student $p_s^{\tau}$ \cite{hinton2015distilling}, defined as
\begin{small} 
\begin{equation}
\setlength{\abovedisplayskip}{1pt}
\setlength{\belowdisplayskip}{1pt}
\begin{aligned}
\mathcal{L}_{KL}(p_t^{\tau},p_s^{\tau})=\frac{1}{K}\sum^{K}_{k=1} p_t^{\tau}(k)log\frac{p_t^{\tau}(k)}{p_s^{\tau}(k)} =\mathcal{L}_{CE}(p_t^{\tau},p_s^{\tau}) - H(p_t^{\tau}),& \\
\end{aligned}
\end{equation}
\end{small}
where $p_t^{\tau}(k)$ and $p_s^{\tau}(k)$ are the $k$-th element of the output vector $p_t^{\tau}$ and $p_s^{\tau}$, respectively. $\mathcal{L}_{CE}(.,.)$ represents the Cross-Entropy loss and $H(\cdot)$ means the entropy value. 
Notably, when $p_t^{\tau}$ stays away from $p_s^{\tau}$ (large distillation gap appears), $p_t^{\tau}$ goes to $y$, then $\mathcal{L}_{KL}(p_t^{\tau},p_s^{\tau})$ will degenerate into $\mathcal{L}_{CE}(y,p_s^{\tau})$ below:
\begin{small} 
\begin{equation}
\setlength{\abovedisplayskip}{1pt}
\setlength{\belowdisplayskip}{1pt}
\begin{aligned}
\lim\limits_{p_t^{\tau}\to y}  \mathcal{L}_{KL}(p_t^{\tau},p_s^{\tau}) =\lim\limits_{p_t^{\tau}\to y} (\mathcal{L}_{CE}(p_t^{\tau},p_s^{\tau}) - H(p_t^{\tau})) =\mathcal{L}_{CE}(y,p_s^{\tau}), &
\end{aligned}
\end{equation}
\end{small}
where $H(y)$ is a constant (\emph{i.e.}, 0) since $y$ is the one-hot label. The gradient of $\mathcal{L}_{KL}$ w.r.t. $z_s$ also has 
\begin{small} 
\begin{equation}
\setlength{\abovedisplayskip}{1pt}
\setlength{\belowdisplayskip}{1pt}
\begin{aligned}
&\lim\limits_{p_t^{\tau}\to y} \frac{\partial \mathcal{L}_{KL}}{\partial z_s} = \lim\limits_{p_t^{\tau}\to y} \frac{1}{\tau}(p_s^{\tau}-p_t^{\tau}) =\frac{1}{\tau}(p_s^{\tau}-y), &
\end{aligned}
\label{expert_grad}
\end{equation}
\end{small}
where the gradient for KL loss increasingly degenerates into that for CE loss, resulting student into the emergency of escaping online KD process.
The results in Fig.\ref{dml_kdcl_gap} Right also confirm the above analysis.
As opposed to that, when $p_t^{\tau}$ goes to $p_s^{\tau}$ (the distillation gap becomes small), $\lim\limits_{p_t^{\tau}\to p_s^{\tau}}  \mathcal{L}_{KL}(p_t^{\tau},p_s^{\tau})=0$, therefore no effective knowledge will be distilled from teacher to student. 

\noindent \textbf{How to yield an appropriate gap?} Inspired by the above, we need to yield an appropriate distillation gap $G$ to ensure that student can always learn effective knowledge from the teacher throughout the training. In other words, the learning pace of student should continuously keep consistent with that of teacher. Otherwise, the online KD process will terminate. To this end, we propose to maintain an appropriate $G$.
When $G$ is larger than a threshold $\delta$, namely \emph{switching threshold}, we terminate teacher and keep only student learn from teacher, such training status is called \emph{expert mode}.  When expert mode progresses, $G$ will decrease until less than $\delta$, it will switch to the other training status of mutually learning between teacher and student, namely \emph{learning mode}. The above two modes alternatively switch under an appropriate $\delta$ to keep improving the student's performance. Next, we will offer the details for learning mode (see Sec.\ref{learning_m}) and expert mode (see Sec.\ref{expert_m}), which pave the way to our proposed adaptive switching threshold $\delta$ (see Sec.\ref{threshold_bound}).

\subsection{Learning Mode: Independent \emph{\textbf{vs}} Reciprocal}
\label{learning_m}
Unlike \cite{hinton2015distilling,Cho_2019_ICCV,jin2019knowledge} that pre-train a teacher network in advance, the goal of learning mode is to reduce the distillation gap by training teacher and student network from scratch. Naturally, one naive strategy is to train the teacher independently with the supervision of one-hot label.
Then the loss function of teacher and student is given as
\begin{small} 
\begin{equation}
\setlength{\abovedisplayskip}{1pt}
\setlength{\belowdisplayskip}{1pt}
\begin{aligned}
&\mathcal{L}_s^l=\mathcal{L}_{CE}(y,p_s^1) +\alpha \tau^2  \mathcal{L}_{KL}(p^{\tau,l}_t,p_s^{\tau}), \quad   \mathcal{L}_t^l=\mathcal{L}_{CE}(y,p^{1,l}_t),
\end{aligned}
\label{learning_teacher_1}
\end{equation}
\end{small} 
where $p^{\tau,l}_t$ and $p_s^{\tau}$ are the predictions of teacher and student, respectively. $\alpha$ is a balancing hyperparameter. 
Unfortunately, the independently trained teacher provides poor improvement for student (see Sec.\ref{ablation_1}).  Inspired by the fact that the teacher can benefit from reciprocal training \cite{zhang2018deep,Guo_2020_CVPR} and, in turn, admit better guidance for student, we propose to reciprocally train student and teacher, therefore $\mathcal{L}_t^l$ in Eqn.(\ref{learning_teacher_1}) can be upgraded to
\begin{small} 
\begin{equation}
\setlength{\abovedisplayskip}{1pt}
\setlength{\belowdisplayskip}{1.5pt}
\begin{aligned}
&\mathcal{L}_t^l=\mathcal{L}_{CE}(y,p^{1,l}_t) + \beta \tau^2 \mathcal{L}_{KL}(p_s^{\tau},p^{\tau,l}_t),
\end{aligned}
\label{learning_loss}
\end{equation}
\end{small} 
where $\beta$ is a balancing hyperparameter. Thus we can compute the gradient of $\mathcal{L}_s^l$ and $\mathcal{L}_t^l$ w.r.t. $z_s$ and $z_t$, \emph{i.e.},
\begin{small} 
\begin{equation}
\setlength{\abovedisplayskip}{1pt}
\setlength{\belowdisplayskip}{1pt}
\label{learning_grad}
\begin{aligned}
&\partial \mathcal{L}_s^l / \partial z_s = (p_s^1-y) + \alpha \tau (p_s^{\tau}-p^{\tau,l}_t), \quad  \partial \mathcal{L}_t^l / \partial z_t = (p^{1,l}_t-y) + \beta \tau (p^{\tau,l}_t-p_s^{\tau}).
\end{aligned}
\end{equation}
\end{small} 
In learning mode, the teacher usually converges faster (yield higher accuracy), owing to its superior learning ability, therefore \emph{the distillation gap will increasingly grow as the training progresses. Meanwhile, for the student, KL loss exhibits a trend to be functionally equivalent to CE loss, causing the effect of knowledge distillation to be weakened}. In this case, SwitOKD will switch to expert mode.

\subsection{Expert Mode: Turn to Wait for Student}
\label{expert_m}
To mitigate the adversarial impact of large distillation gap on student, SwitOKD attempts to pause the training of teacher while make student continuously learn from teacher, to keep the learning pace consistent, that sets it apart from previous online distillation methods \cite{zhang2018deep,Guo_2020_CVPR}. Indeed, a teacher that is suitable for student rather than one who perfectly imitates one-hot label, can often improve student's performance, in line with our view of an appropriate distillation gap.
Accordingly, the loss function for student is similar in spirit to that of Eqn.(\ref{learning_teacher_1}):
\begin{small} 
\begin{equation}
\setlength{\abovedisplayskip}{1pt}
\setlength{\belowdisplayskip}{1pt}
\begin{aligned}
&\mathcal{L}_s^e=\mathcal{L}_{CE}(y,p_s^1) + \alpha \tau^2 \mathcal{L}_{KL}(p^{\tau,e}_t,p_s^{\tau}),&
\end{aligned}
\label{expert_loss}
\end{equation}
\end{small} 
where $p^{\tau,e}_t$ is the prediction of teacher network under expert mode.
Thus the gradient of $\mathcal{L}_s^e$ w.r.t. $z_s$ is computed as
\begin{small} 
\begin{equation}
\setlength{\abovedisplayskip}{1pt}
\setlength{\belowdisplayskip}{1pt}
\begin{aligned}
&\partial \mathcal{L}_s^e / \partial z_s = (p_s^1-y) + \alpha \tau (p_s^{\tau}-p^{\tau,e}_t). &
\end{aligned}
\label{expert_grad}
\end{equation}
\end{small} 
In such mode, the student will catch up or even surpass teacher as the training progresses, resulting into no effective knowledge distilled from teacher to student. Then SwitOKD will switch back to learning mode based on our adaptive switching threshold. We discuss that in the next section.

\subsection{Adaptive Switching Threshold: Extending Online Knowledge Distillation’s Lifetime}
\label{threshold_bound} 
Intuitively, a naive strategy is to \emph{manually} select a \emph{fixed} value of $\delta$, which, however, is inflexible and difficult to yield an appropriate distillation gap for improving the student (see Sec.\ref{ablation_1}).
We propose an adaptive switching threshold for $\delta$, which offer insights into how to \emph{automatically} switch between learning mode and expert mode.
First, observing that the distillation gap $G=||p_s^{\tau}-p_t^{\tau}||_1 < ||p_s^{\tau}-y||_1$ on average because the teacher is superior to student, and 
\begin{small} 
\begin{equation}
\setlength{\abovedisplayskip}{1.5pt}
\setlength{\belowdisplayskip}{4pt}
\begin{aligned}
||p_s^{\tau}-p_t^{\tau}||_1&=||(p_s^{\tau}-y)-(p_t^{\tau}-y)||_1  \geq ||p_s^{\tau}-y||_1-||p_t^{\tau}-y||_1, 
\end{aligned}
\end{equation}
\end{small}
which further yields $||p_s^{\tau}-y||_1-||p_t^{\tau}-y||_1\leq G<||p_s^{\tau}-y||_1$, leading to
\begin{small} 
\begin{equation}
\setlength{\abovedisplayskip}{1pt}
\setlength{\belowdisplayskip}{1pt}
\begin{aligned}
\underbrace{||p_s^{\tau}-y||_1-||p_t^{\tau}-y||_1}_{\textbf{lower \ bound}} \leq \delta < \underbrace{||p_s^{\tau}-y||_1}_{\textbf{upper bound}}, &
\end{aligned}
\label{bound_delta}
\end{equation}
\end{small}
which, as aforementioned in Sec.\ref{introduction}, ought to be neither too large nor too small.
To this end, we propose to adaptively adjust $\delta$. Based on Eqn.(\ref{bound_delta}), we can further reformulate $\delta$ to be: 
\begin{small} 
\begin{equation}
\setlength{\abovedisplayskip}{3pt}
\setlength{\belowdisplayskip}{3pt}
\begin{aligned}
\delta=||p_s^{\tau}-y||_1-\varepsilon||p_t^{\tau}-y||_1,  0< \varepsilon \leq1.&
\end{aligned}
\label{epsilon}
\end{equation}
\end{small}
It is apparent that $\varepsilon$ approaching either 1 or 0 is equivalent to lower or upper bound of Eqn.(\ref{bound_delta}). 
Unpacking Eqn.(\ref{epsilon}), the effect of $\varepsilon$ is expected to be: when $G$ becomes large, $\delta$ will be decreased towards $||p_s^{\tau}-y||_1-||p_t^{\tau}-y||_1$ provided $\varepsilon$ approaching $1$, then $G>\delta$ holds, which naturally enters into expert mode, and switches back into learning mode vice versa; see Fig.\ref{overview}(b).

\noindent \textbf{Discussion on} $\bm{\varepsilon}$. As per Eqn.(\ref{epsilon}), the value of $\delta$ closely relies on $\varepsilon$, which actually plays the role of tracking the changing trend of $G$. Intuitively, once the teacher learns faster than student, $G$ will be larger, while $||p_t-y||_1 < ||p_s-y||_1$ holds from Eq.(\ref{epsilon}). Under such case, small value of $\delta$ is expected, leading to a larger value of $\varepsilon$, and vice versa. Hence, $\varepsilon$ is inversely proportional to $r = \frac{||p_t^{\tau}-y||_1}{||p_s^{\tau}-y||_1}$.  However, if $G$ is very large, the student cannot catch up with the teacher; worse still, the teacher is constantly paused (trapped in expert mode) and cannot improve itself to distill knowledge to student, \emph{making the online KD process terminated}. Hence, we decrease $\delta$, so that, observing that $p_t$ and $p_s$ are very close during the early training time, the teacher can pause more times initially to make student to be in line with teacher, to avoid being largely fall behind at the later training stage (see Sec.\ref{why_work} for detailed validations). Following this, we further decrease the value of $r$, such that $r = \frac{||p_t^{\tau}-y||_1}{||p_s^{\tau}-y||_1 + ||p_t^{\tau}-y||_1}$, to balance learning mode and expert mode. 
For normalization issue, we reformulate $\varepsilon = e^{-r}$, leading to the final adaptive switching threshold $\delta$ to be: 
\begin{small} 
\begin{equation}
\setlength{\abovedisplayskip}{4pt}
\setlength{\belowdisplayskip}{0.3pt}
\begin{aligned}
\boxed{ \delta=||p_s^{\tau}-y||_1-e^{-\frac{||p_t^{\tau}-y||_1}{||p_s^{\tau}-y||_1 + ||p_t^{\tau}-y||_1}} ||p_t^{\tau}-y||_1.} &
\end{aligned}
\label{final_delta}
\end{equation}
\end{small}

Unlike the existing arts \cite{zhang2018deep,Guo_2020_CVPR}, where they fail to focus on student to follow the principle of KD, SwitOKD can \emph{extend online knowledge distillation's lifetime}, and thus largely improve student's performance, while keep our teacher be basically on par with theirs, thanks to Eqn.(\ref{final_delta}); see Sec.\ref{com_with_others} for validations.

\subsection{Optimization}  
\label{optim}
The above specifies the adaptive switching strategy between two training modes.
Specifically, we kick off SwitOKD with learning mode to minimize $\mathcal{L}_s^l$ and $\mathcal{L}_t^l$, then the training mode is switched into expert mode to minimize $\mathcal{L}_s^e$ when $G>\delta$. Following that, SwitOKD switches back to learning mode when $G\le \delta$.
The whole training process is summarized in Algorithm \ref{alg_algorithm}.

\subsection{Multi-network Learning Framework}  
\label{multi_n}

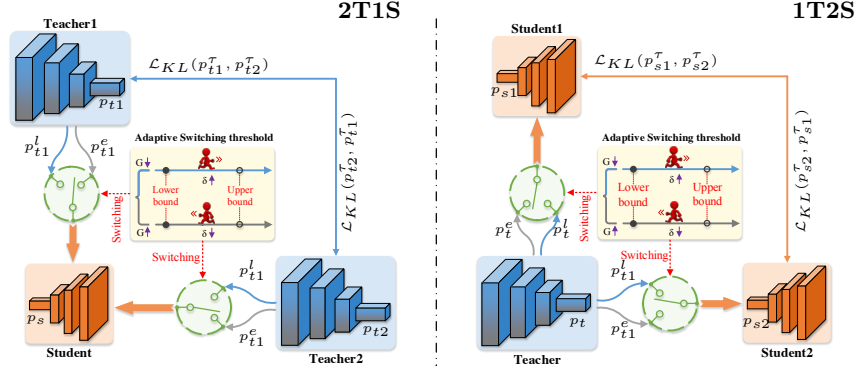
\begin{figure*}[t]
\centering
\setlength{\abovecaptionskip}{0.1cm}
\setlength{\belowcaptionskip}{-0.45cm}

\begin{overpic}[width=0.9\textwidth]{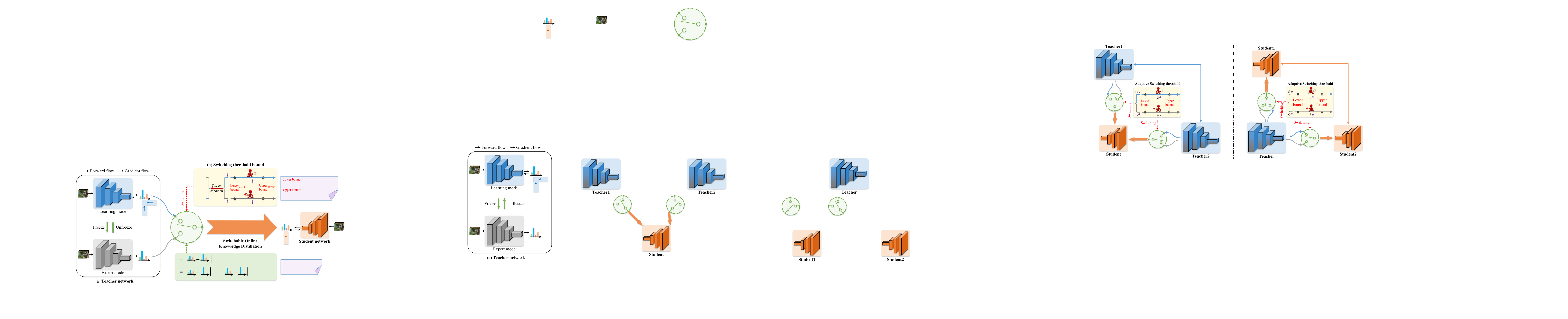}

\put(2,26){\tiny{$p_{t1}^l$}}
\put(10,26){\tiny{$p_{t1}^e$}}
\put(2.3,5.3){\tiny{$p_s$}}

\put(11,31.5){\tiny{$p_{t1}$}}
\put(43,4){\tiny{$p_{t2}$}}
\put(17,36){\tiny{$\mathcal{L}_{KL}(p_{t1}^{\tau},p_{t2}^{\tau}$)}}
\put(39.5,16){\tiny{ \rotatebox{90}{$\mathcal{L}_{KL}(p_{t2}^{\tau},p_{t1}^{\tau}$)}} }

\put(28,11){\tiny{$p_{t1}^l$}}
\put(28,3){\tiny{$p_{t1}^e$}}

\put(40,42){\textbf{2T1S}}

\put(59,16){\tiny{$p_{t}^e$}}
\put(66,16){\tiny{$p_{t}^l$}}
\put(68,5){\tiny{$p_t$}}

\put(58.5,33){\tiny{$p_{s1}$}}
\put(89,5){\tiny{$p_{s2}$}}
\put(71,36.5){\tiny{$\mathcal{L}_{KL}(p_{s1}^{\tau},p_{s2}^{\tau}$)}}
\put(94.5,16){\tiny{ \rotatebox{90}{$\mathcal{L}_{KL}(p_{s2}^{\tau},p_{s1}^{\tau}$)}} }

\put(73,11){\tiny{$p_{t1}^l$}}
\put(73,4){\tiny{$p_{t1}^e$}}

\put(95,42){\textbf{1T2S}}

\end{overpic}

\caption{The multi-network framework for training 3 networks simultaneously, including two fundamental  basis topologies: \textbf{2T1S} (\textbf{Left}) and \textbf{1T2S} (\textbf{Right}).  }
\label{multi_net}
\end{figure*}

To endow SwitOKD with the extendibility to multi-network setting with large distillation gap, we divide these networks into multiple teachers and students, involving switchable online distillation between teachers and students, which is built by two types of fundamental basis topologies below: multiple teachers \emph{vs} one student and one teacher \emph{vs} multiple students.
For ease of understanding, we take 3 networks as an example and denote the basis topologies as \textbf{2T1S} and \textbf{1T2S}, respectively; see Fig.\ref{multi_net}.
Notably, the training between each teacher-student pair directly follows SwitOKD, while two teachers for \textbf{2T1S} (or two students for \textbf{1T2S}) mutually transfer knowledge in a conventional two-way manner. Note that, for \textbf{1T2S}, only when the switching conditions between teacher and both students are triggered, will the teacher be completely suspended. The detailed validation results are reported in Sec.\ref{multi_nets_result}.

\begin{algorithm}[t]
\caption{SwitOKD: Switchable Online Knowledge Distillation}
\label{alg_algorithm}  
\begin{multicols}{2}  
\textbf{Input}: learning rate $\eta_1$, $\eta_2$, student network $\mathcal{S}$ parameterized by $\theta_{s}$, teacher network $\mathcal{T}$ parameterized by $\theta_{t}$  \\
\textbf{Output}: Trained $\mathcal{S}$, $\mathcal{T}$
\begin{algorithmic}[1] 
\STATE Randomly initialize $\mathcal{S}$ and $\mathcal{T}$.
\FOR{number of training iterations}
\STATE Compute $G=||p_s^{\tau}-p_t^{\tau}||_1$.
\STATE Compute $\delta$ by Eqn.(\ref{final_delta}).
\IF {$G\leq \delta$}
\STATE \textbf{\# Learning Mode}
\STATE Estimate $\mathcal{L}_s^l$, $\mathcal{L}_t^l$ with Eqn.(\ref{learning_loss}).
\STATE Update $\theta_{s}$, $\theta_{t}$:
\STATE \qquad $\theta_{s} \gets \theta_{s} - \eta_1 \frac{\partial \mathcal{L}_s^l}{\partial {z_s}}   \frac{\partial z_s}{\partial {\theta_{s}}}$
\STATE \qquad $\theta_{t} \gets \theta_{t} - \eta_2 \frac{\partial \mathcal{L}_t^l}{\partial {z_t}}  \frac{\partial z_t}{\partial {\theta_{t}}}$
\ELSE
\STATE \textbf{\# Expert Mode}
\STATE Estimate $\mathcal{L}_s^e$ with Eqn.(\ref{expert_loss}).
\STATE Freeze $\theta_{t}$, update $\theta_{s}$:
\STATE \qquad $\theta_{s} \gets \theta_{s} - \eta_1 \frac{\partial \mathcal{L}_s^e}{\partial {z_s}}   \frac{\partial z_s}{\partial {\theta_{s}}}$
\ENDIF
\ENDFOR
\end{algorithmic}
\end{multicols}
\end{algorithm}

%% file: experiment.tex
\section{Experiment}
To validate the effectiveness of SwitOKD, we experimentally evaluate various state-of-the-art backbone networks via student-teacher pair below: MobileNet \cite{howard2017mobilenets}, MobileNetV2 \cite{sandler2018mobilenetv2} (sMobileNetV2 means the width multiplier is s), ResNet \cite{he2016deep} and Wide ResNet (WRN) \cite{zagoruyko2016wide} over the  typical image classification datasets:
\textbf{CIFAR-10} and \textbf{CIFAR-100} \cite{krizhevsky2009learning} are natural image datasets, including 32$\times$32 RGB images containing 10 and 100 classes. Both of them are split into a training set with 50k images and a test set with 10k images.
\textbf{Tiny-ImageNet} \cite{le2015tiny} consists of 64$\times$64 color images from 200 classes. Each class has 500 training images, 50 validation images, and 50 test images.
\textbf{ImageNet} \cite{russakovsky2015imagenet} contains 1k object classes with  about 1.2 million images for training and 50k images for validation.

\begin{figure}[t]
\centering
\setlength{\abovecaptionskip}{0.1cm}
\setlength{\belowcaptionskip}{-0.53cm}
\includegraphics[width=0.78\textwidth]{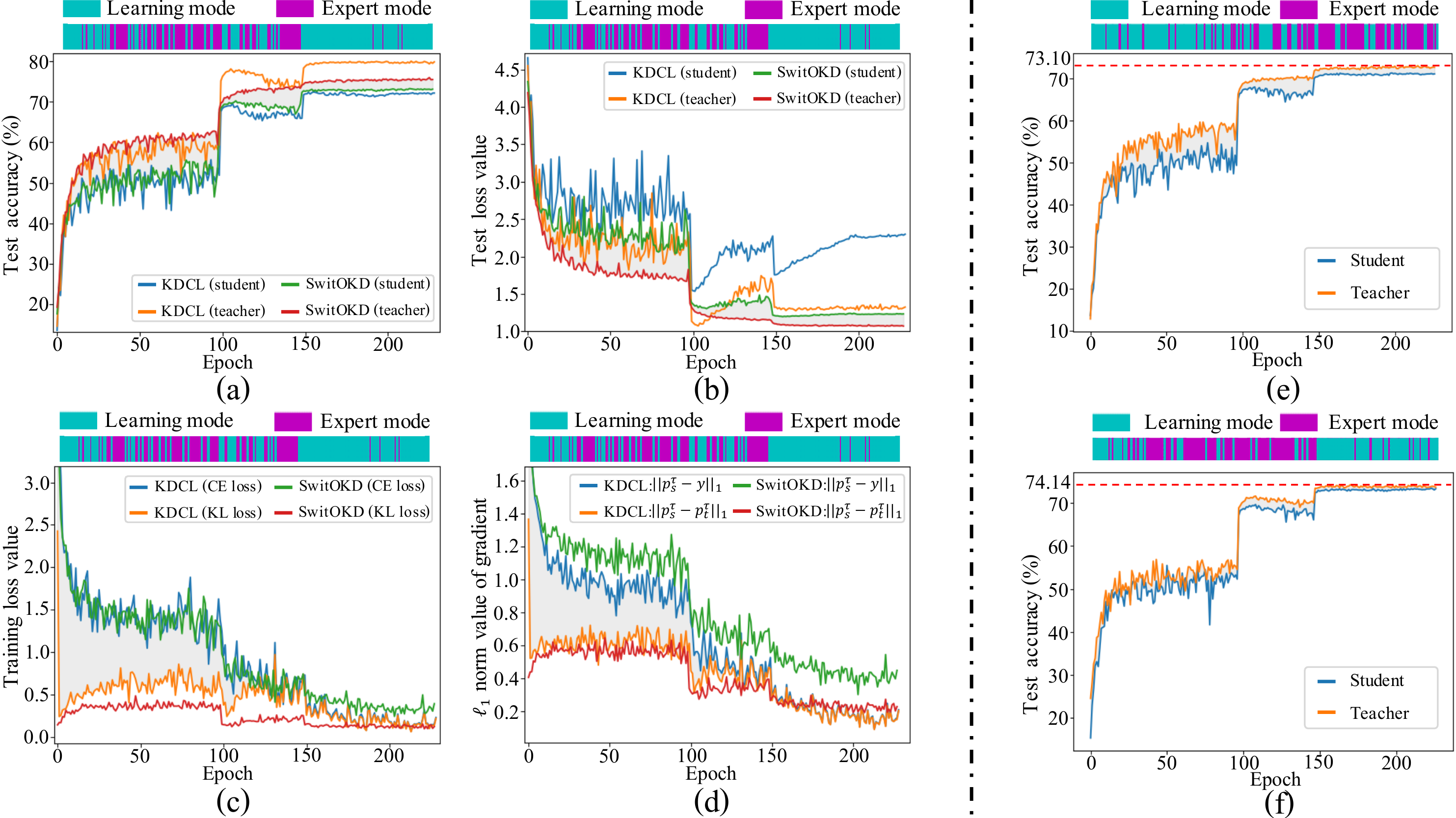}

\caption{\textbf{Left:} Illustration of test accuracy (a) and loss (b) for SwitOKD and KDCL \cite{Guo_2020_CVPR}. From the perspective of student, (c) shows the comparison of CE loss and KL loss, while (d) is $\ell_1$ norm value of the gradient for CE loss and KL loss. \textbf{Right:} Illustration of why the parameter $r$ should be decreased from $r = \frac{||p_t^{\tau}-y||_1}{||p_s^{\tau}-y||_1}$ (e) to $r = \frac{||p_t^{\tau}-y||_1}{||p_s^{\tau}-y||_1 + ||p_t^{\tau}-y||_1}$ (f).
The color bar shows the switching process of SwitOKD, where the \textcolor[RGB]{86,188,189}{cyan} and the \textcolor[RGB]{174,36,185}{magenta} denote learning mode and expert mode, respectively. }
\end{figure}

\subsection{Experimental Setup}
We implement all networks and training procedures with pytorch \cite{paszke2017automatic} on an NVIDIA GeForce GTX 1080 Ti GPU and an Intel(R) Core(TM) i7-6950X CPU @ 3.00GHz. For \textbf{CIFAR-10/100}, we use Adam optimizer with the momentum of 0.9, weight decay of 1e-4 and set batch size to 128. The initial learning rate is 0.01 and then divided by 10 at 140, 200 and 250 of the total 300 epochs. For \textbf{Tiny-ImageNet} and \textbf{ImageNet}, we adopt SGD as the optimizer, and set momentum to 0.9 and weight decay to 5e-4. Specifically, for Tiny-ImageNet, we set batch size to 128, the initial learning rate to 0.01, and the learning rate is dropped by 0.1 at 100, 140 and 180 of the total 200 epochs. For ImageNet, batch size is 512, while the initial learning rate is 0.1 (dropped by 0.1 every 30 epochs and trained for 120 epochs). As for the hyperparameter, we set $\alpha$, $\beta$ and $\tau$ to 1, and $\tau$=\{2,3\} for the classic distillation \cite{hinton2015distilling,Guo_2020_CVPR}.

Previous sections (Sec.\ref{learning_m}, \ref{expert_m}, \ref{threshold_bound} and \ref{multi_n}) explicate how adaptive switching strategy benefits a student. We offer practical insights into why SwitOKD works well, including ablation study and comparison with the state-of-the-arts, as well as extendibility to multiple networks.

\subsection{Why does SwitOKD Work Well?}
\label{why_work}
One of our aims is to confirm that the core idea of our SwitOKD --- using an adaptive switching threshold to achieve adaptive switching strategy --- can possess an appropriate distillation gap for the improvement of student. The other is to verify why the parameter $r$ (Sec.\ref{threshold_bound}) should be decreased.  We perform online distillation with a compact student network (ResNet-32) and a powerful teacher network (WRN-16-8 and WRN-16-2) on CIFAR-100.

Fig.4(a)(b) illustrate that the performance of student is continuously improved with smaller accuracy gap (\textcolor[RGB]{123,123,123}{gray} area) compared to KDCL, confirming that our switching strategy can effectively calibrate the distillation gap to extend online KD's lifetime, in keeping with the core idea of SwitOKD.
As an extension of Fig.\ref{dml_kdcl_gap}, Fig.4(c) reveals that KL loss for SwitOKD keeps far away from CE loss throughout the training unlike KDCL.
Akin to that, the gradient for KL loss $||p_s^{\tau}-p_t^{\tau}||_1$ (refer to Eqn.(\ref{gap_quantify})) keeps divergent from that for CE loss $||p_s^{\tau}-y||_1$; see Fig.4(d).
Especially, \emph{the color bar} illustrates the process of switching two modes on top of each other: when to pause the teacher --- expert mode (\textcolor[RGB]{174,36,185}{magenta}), or restart training --- learning mode (\textcolor[RGB]{86,188,189}{cyan}), reflecting that an appropriate gap holds with adaptive switching operation.

Fig.4(e)(f)  validate the findings below: when $r = \frac{||p_t^{\tau}-y||_1}{||p_s^{\tau}-y||_1}$ (e), the teacher is rarely paused at the early stage of training, then the student largely falls behind at the later stage, leading to poor teacher (73.10\% \emph{vs} 74.14\%) and student (71.89\% \emph{vs} 73.47\%), confirming our analysis in Sec.\ref{threshold_bound} --- $r = \frac{||p_t^{\tau}-y||_1}{||p_s^{\tau}-y||_1 + ||p_t^{\tau}-y||_1}$ (f) is desirable to balance learning mode and expert mode.

\begin{wraptable}{r}{6.1cm}
\centering
\setlength{\abovecaptionskip}{-0.02cm}
\setlength{\belowcaptionskip}{-0.9cm}
\caption{Ablation study about the effectiveness of each component, of which constitutes SwitOKD. The best results are reported with \textbf{boldface}.}
\scriptsize
\begin{tabular}{c|c|c|c|c}
\toprule
Case &\tabincell{c}{Threshold \\ $\delta$}     &\tabincell{c}{Switching \\ or not}  &\tabincell{c}{Teacher's \\ loss $\mathcal{L}_t^l$}     &CIFAR-100     \\
\hline\hline
\cellcolor{mygreen}\textbf{A} &\cellcolor{mygreen}-                                        &\cellcolor{mygreen}\ding{55}                                &\cellcolor{mygreen}Eqn.(\ref{learning_loss})          &\cellcolor{mygreen}{72.91}      \\
\hline
\multirow{3}{*}{\textbf{B}}
&$\delta=0.2$                     &                                              &                                                 &{72.92}      \\
&$\delta=0.6$                     &\ding{51}                                &Eqn.(\ref{learning_loss})          &{72.83}      \\
&$\delta=0.8$                     &                                              &                                                 &{73.00}      \\
\hline
\cellcolor{mygreen}\textbf{C} &\cellcolor{mygreen}Eqn.(\ref{final_delta})       &\cellcolor{mygreen}\ding{51}                                &\cellcolor{mygreen}Eqn.(\ref{learning_teacher_1})          &\cellcolor{mygreen}{72.72}      \\
\hline
\textbf{D} &Eqn.(\ref{final_delta})       &\ding{51}                                &Eqn.(\ref{learning_loss})          &\textbf{73.47}      \\
\bottomrule
\end{tabular}
\label{ablation_study}
\end{wraptable}

\subsection{Ablation Studies}
\subsubsection{Is Each Component of SwitOKD Essential?}
\label{ablation_1}
To verify the effectiveness of several components constituting SwitOKD --- \emph{switching strategy}, \emph{adaptive switching threshold} and \emph{teacher's training strategy},
we construct ablation experiments with ResNet-32 (student) and WRN-16-2 (teacher) from the following cases: \textbf{A}: SwitOKD without switching; \textbf{B}: SwitOKD with fixed $\delta$ (\emph{i.e.}, $\delta \in \{0.2, 0.6, 0.8\}$); \textbf{C}: teacher's loss $\mathcal{L}_t^l$ (Eqn.(\ref{learning_loss}) \emph{vs} Eqn.(\ref{learning_teacher_1})); \textbf{D}: the proposed SwitOKD.
Table \ref{ablation_study} summarizes our findings, which suggests that SwitOKD shows great superiority (73.47\%) to other cases. Especially for case \textbf{B}, the manual $\delta$ fails to yield an appropriate distillation gap for improving the performance of  student, confirming the importance of adaptive switching threshold, subject to our analysis (Sec.\ref{threshold_bound}). Notably, the student for case \textbf{C} suffers from a large accuracy loss, verifying the benefits of reciprocal training on improving the performance of student (Sec.\ref{learning_m}).

\begin{wraptable}{r}{6.1cm}
\centering
\setlength{\abovecaptionskip}{-0.02cm}
\setlength{\belowcaptionskip}{-0.8cm}
\caption{Ablation study about the effectiveness of varied temperature $\tau$ on CIFAR-100. The best results are reported with \textbf{boldface}.}
\scriptsize
\begin{tabular}{c|c|c|c|c|c|c}
\toprule
$\tau$                                         &0.5       &1           &2   &5  &8  &10    \\
\hline\hline
\textbf{SwitOKD}                        &64.95   &67.24     &\textbf{67.80}   &66.30 &66.00  &65.64  \\
\bottomrule
\end{tabular}

\label{ablation_study_tau}
\end{wraptable}

\subsubsection{Why does the Temperature \textbf{$\tau$} Benefit SwitOKD?}
The temperature $\tau$ \cite{hinton2015distilling} usually serves as a factor to smooth the predictions of student and teacher. Empirically, temperature parameter enables the output of student to be closer to that of teacher (and thus reduce the distillation gap), improving the performance, in line with our perspective in Sec.\ref{expert_m}. To highlight the effectiveness of SwitOKD, we simply set $\tau=1$ for our experiments. To further verify the effectiveness of varied $\tau \in \{0.5,1,2,5,8,10\}$, we perform the ablation experiments with MobileNetV2 (student) and WRN-16-2 (teacher). 
Table \ref{ablation_study_tau} summarizes the findings. The optimal student (67.80\%) is achieved with a slightly higher $\tau^*=2$, implying that $\tau$ contributes to the calibration of the distillation gap. Note that when $\tau=10$, the accuracy of student rapidly declines, implying that excessive smoothness can make the gap beyond an optimal range and, in turn, harm the performance of student, consistent with our view of an appropriate gap in Sec.\ref{tradition}.

\begin{table*}[t]
\centering
\setlength{\belowcaptionskip}{0.1cm}
\setlength{\abovecaptionskip}{-0.02cm}
\caption{Accuracy (\%) comparison on Tiny-ImageNet and CIFAR-10/100. (.M) denotes the number of parameters. All the values are measured by computing mean and standard deviation across 3 trials with random seeds. The best results are reported with \textbf{boldface}.}

\tiny

\begin{tabular}{cc|c|c|c|c}
\toprule
\multicolumn{2}{c|}{Backbone} &
\multicolumn{1}{c|}{Vanilla} &

\multicolumn{1}{c|}{DML \cite{zhang2018deep}} &
\multicolumn{1}{c|}{KDCL \cite{Guo_2020_CVPR}}

&    \multicolumn{1}{c}{\textbf{SwitOKD}}  \\

\hline\hline
\multicolumn{6}{l}{\rule{0pt}{7pt} \textbf{Tiny-ImageNet}} \\
\hline

 \cellcolor{mygreen}\quad\quad Student        &\cellcolor{mygreen}1.4MobileNetV2(4.7M)    &\cellcolor{mygreen}50.98{\tiny{$\pm$0.32}}        &\cellcolor{mygreen}55.70{\tiny{$\pm$0.61}}     &\cellcolor{mygreen}57.79{\tiny{$\pm$0.30}}     &\cellcolor{mygreen}\textbf{58.71\tiny{$\pm$0.11}}    \\

\quad\quad Teacher                               &ResNet-34(21.3M)    &63.18{\tiny{$\pm$0.37}}          &64.49{\tiny{$\pm$0.43}}     &\textbf{65.47{\tiny{$\pm$0.32}}}   &{63.31\tiny{$\pm$0.04}}    \\

\hline
 \cellcolor{mygreen}\quad\quad Student        &\cellcolor{mygreen}ResNet-20(0.28M)    &\cellcolor{mygreen}52.35{\tiny{$\pm$0.15}}       &\cellcolor{mygreen}53.98{\tiny{$\pm$0.26}}     &\cellcolor{mygreen}53.74{\tiny{$\pm$0.39}}     &\cellcolor{mygreen}\textbf{55.03\tiny{$\pm$0.19}}    \\

\quad\quad Teacher                               &WRN-16-2(0.72M)    &56.59{\tiny{$\pm$0.22}}            &57.45{\tiny{$\pm$0.19}}     &\textbf{57.71{\tiny{$\pm$0.30}}}   &{57.41}\tiny{$\pm$0.06}   \\

\hline\hline
\multicolumn{6}{l}{\rule{0pt}{7pt} \textbf{CIFAR-10}}  \\
\hline
\cellcolor{mygreen}\quad\quad Student      &\cellcolor{mygreen}WRN-16-1(0.18M)   &\cellcolor{mygreen}91.45{\tiny{$\pm$0.06}}         &\cellcolor{mygreen}91.96{\tiny{$\pm$0.08}} &\cellcolor{mygreen}91.86{\tiny{$\pm$0.11}}       &\cellcolor{mygreen}\textbf{92.50\tiny{$\pm$0.17}}    \\

\quad\quad Teacher                      &WRN-16-8(11.0M)   &95.21\tiny{$\pm$0.12}       &95.06{\tiny{$\pm$0.05}}  &\textbf{95.33{\tiny{$\pm$0.17}}}   &{94.76\tiny{$\pm$0.12}}    \\

\hline\hline
\multicolumn{6}{l}{\rule{0pt}{7pt} \textbf{CIFAR-100}} \\
\hline

\cellcolor{mygreen}\quad\quad Student       &\cellcolor{mygreen}\quad\quad0.5MobileNetV2(0.81M)$\quad\quad$        &\cellcolor{mygreen}{\quad 60.07\tiny{$\pm$0.40}\quad }                &\cellcolor{mygreen}{\quad 66.23\tiny{$\pm$0.36}\quad }   &\cellcolor{mygreen}{\quad 66.83\tiny{$\pm$0.05}\quad }    &\cellcolor{mygreen}\textbf{\quad 67.24\tiny{$\pm$0.04}\quad }     \\

\quad\quad Teacher                       &WRN-16-2(0.70M)      &72.90\tiny{$\pm$0.09}            &73.85\tiny{$\pm$0.21}    &{73.75\tiny{$\pm$0.26}}    &\textbf{73.90\tiny{$\pm$0.40}}     \\

\bottomrule
\end{tabular}

\label{result_cifar}
\end{table*}

\begin{figure}[t]
\centering
\setlength{\abovecaptionskip}{0.1cm}
\setlength{\belowcaptionskip}{-0.6cm}
\includegraphics[width=0.85\textwidth]{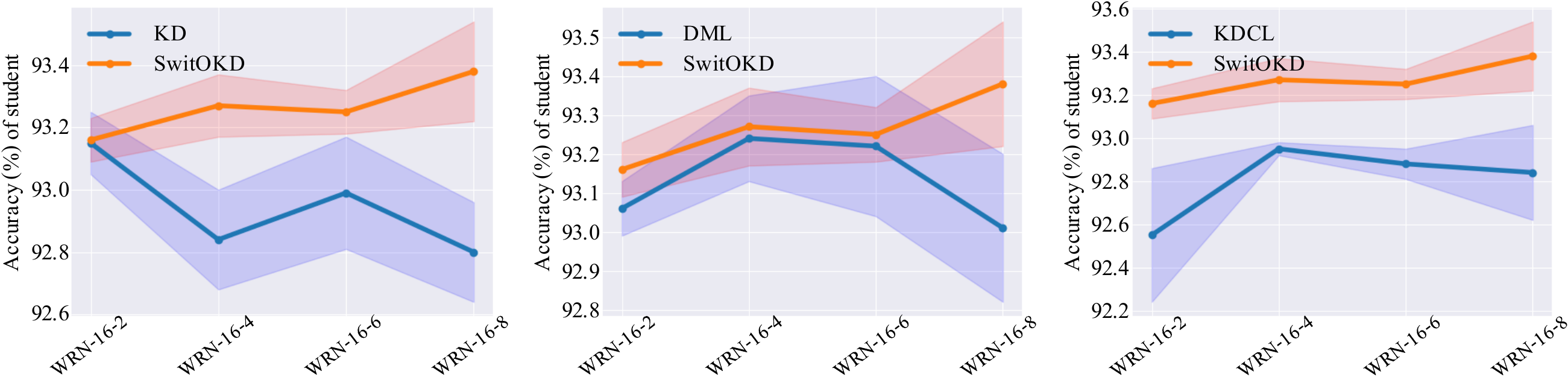}
\caption{Illustration of how an appropriate distillation gap yields better student. For KD, DML and KDCL, the accuracy of student (ResNet-20) rapidly declines as the teacher turns to higher capacity (WRN-16-2 to WRN-16-8). As opposed to that, SwitOKD grows steadily, owing to an appropriate distillation gap.  }
\label{gap_increase}
\end{figure}

\subsection{Comparison with Other Approaches}
\label{com_with_others}
To verify the superiority of SwitOKD, we first compare with typical online KD methods, including: 1) DML \cite{zhang2018deep} is equivalent to learning mode for SwitOKD;
2) KDCL \cite{Guo_2020_CVPR} studies the effect of large accuracy gap at the test phase on online distillation process, but they pay more attention to teacher instead of student.
For brevity,  ``vanilla" refers to the backbone network trained from scratch with classification loss alone. 
A compact student and a powerful teacher constitute the student-teacher network pair with \emph{large distillation gap} at the training phase.

\begin{table*}[t]
\centering
\setlength{\belowcaptionskip}{0.4cm}
\setlength{\abovecaptionskip}{-0.02cm}
\caption{Accuracy (\%) comparison of student network with offline KD methods (seen as expert mode of SwitOKD) on CIFAR-100. (.M) denotes the number of parameters. The best results are reported with \textbf{boldface}.}
\scriptsize
\begin{tabular}{cc|c|c|c|c|c|c|c}
\toprule
\multicolumn{2}{c|}{Backbone} &
\multicolumn{1}{c|}{Vanilla} &
\multicolumn{1}{c|}{KD \cite{hinton2015distilling}} &
\multicolumn{1}{c|}{FitNet \cite{romero2014fitnets}} &
\multicolumn{1}{c|}{AT \cite{zagoruyko2016paying}}&
\multicolumn{1}{c|}{CRD \cite{tian2019contrastive} } &
\multicolumn{1}{c|}{RCO \cite{jin2019knowledge} }

&    \multicolumn{1}{c}{\textbf{SwitOKD}}  \\
\hline\hline
\rowcolor{mygreen}

Student      &WRN-16-2(0.70M)   &72.79   &74.49    &73.44   &73.35    &75.01  &75.36  &\textbf{75.95}   \\

 Teacher    &WRN-40-2(2.26M)    &76.17   &- &-   &-  &-   &-  &\textbf{76.54}  \\

%
\bottomrule
\end{tabular}
\label{sota}
\end{table*}

\begin{table}[t]
\centering
\setlength{\belowcaptionskip}{0.1cm}
\setlength{\abovecaptionskip}{-0.02cm}
\caption{Top-1 accuracy (\%) on ImageNet dataset. (.M) denotes the number of parameters.  The best results are reported with \textbf{boldface}.}
\tiny
\begin{tabular}{cc|c|c|c|c}
\toprule
\multicolumn{2}{c|}{Backbone}         &\quad Vanilla$\quad$    &\quad DML \cite{zhang2018deep}$\quad$       &\quad KDCL \cite{Guo_2020_CVPR}$\quad$        &\quad\textbf{SwitOKD}$\quad$          \\
\hline\hline
\rowcolor{mygreen}
Student      &ResNet-18(11.7M)       &69.76      &70.81                                          &70.91                    &\textbf{71.75}      \\
Teacher      &ResNet-34(21.8M)       &73.27      &73.47                                          &\textbf{73.70}        &73.65      \\
\hline
\rowcolor{mygreen}
\quad Student$\quad$      &\quad0.5MobileNetV2(1.97M)$\quad$       &63.54       &64.22                                      &63.92                     &\textbf{65.11}      \\
Teacher      &ResNet-18(11.7M)          &69.76       &68.30                                           &\textbf{70.60}            &68.08       \\
\bottomrule
\end{tabular}
\label{comparison_imagenet}
\end{table}

Table \ref{result_cifar} and Fig.\ref{gap_increase} summarize our findings below: 
\emph{First}, switchable online distillation offers a significant and consistent performance improvement over the baseline (vanilla) and the state-of-the-arts for \emph{student}, in line with the principle of KD process. Impressively, SwitOKD achieves 1.05\% accuracy improvement to vanilla on CIFAR-10 (7.17\% on CIFAR-100). Besides, SwitOKD also shows 0.54\% and 0.54\% (WRN-16-1/WRN-16-8) accuracy gain over DML and KDCL on CIFAR-100, respectively. Especially with 1.4MobileNetV2/ResNet-34, SwitOKD still obtains significant performance gains of 7.73\%, 3.01\% and 0.92\% (the gains are substantial for Tiny-ImageNet) over vanilla, DML and KDCL. 

\emph{Second}, our teachers still benefit from SwitOKD and obtain accuracy improvement \emph{basically on a par} with DML and KDCL, confirming our analysis about the adaptive switching threshold $\delta$ (see Eqn.({\ref{final_delta}})) --- balance of learning mode and expert mode. Note that, with 0.5MobileNetV2 and WRN-16-2 on CIFAR-100, our teacher (73.90\%) upgrades beyond the vanilla (72.90\%), even yields comparable accuracy gain (0.05\% and 0.15\%) over DML and KDCL. By contrast, KDCL has most of the best teachers, but with poor students, owing to its concentration on teacher only.

\emph{Finally}, we also validate the effectiveness of SwitOKD for student even under a small distillation gap on CIFAR-10 (see Fig.\ref{gap_increase}), where the students (ResNet-20) still possess significant performance advantages,
confirming the necessity of adaptively calibrating an appropriate gap with adaptive switching threshold $\delta$ in Sec.\ref{threshold_bound}. Especially for Fig.\ref{gap_increase} (b)(c),  as the teacher turns to higher capacity (WRN-16-2 to WRN-16-8), students' accuracy from DML and KDCL rises at the beginning, then rapidly declines, and reaches the best results when the teacher is WRN-16-4. This, in turn, keeps consistent with our analysis (Sec.\ref{tradition}) --- an appropriate distillation gap admits student's improvement.

To further validate the switching strategy between two modes, we also compare SwitOKD with offline knowledge distillation approaches (seen as expert mode of SwitOKD) including KD \cite{hinton2015distilling}, FitNet \cite{romero2014fitnets}, AT \cite{zagoruyko2016paying} and CRD \cite{tian2019contrastive} that require a fixed and pre-trained teacher. Especially, RCO \cite{jin2019knowledge} is similar to our approach, which maintains a reasonable performance gap by manually selecting a series of pre-trained intermediate teachers.    
Table \ref{sota} reveals that SwitOKD achieves superior performance over offline fashions, while exceeds the second best results from RCO by 0.59\%, implying that SwitOKD strictly follows the essential principle of KD with the adaptive switching strategy.

\subsection{Extension to Large-scale Dataset}
Akin to \cite{zhang2018deep,Guo_2020_CVPR}, as a by-product, SwitOKD can effectively be extended to the large-scale dataset (\emph{i.e.}, ImageNet), benefiting from its good generalization ability; see Table \ref{comparison_imagenet}. It is observed that the students' accuracy is improved by 1.99\% and 1.57\% upon the vanilla, which are substantial for ImageNet, validating the scalability of SwitOKD. Particularly, for ResNet-34, our teacher (73.65\%)  outperforms the vanilla (73.27\%) and DML (73.47\%), highlighting the importance of our adaptive switching strategy upon $\delta$ to balance the teacher and student.  Another evidence is shown for 0.5MobileNetV2 and ResNet-18 with larger distillation gap, our student outperforms DML and KDCL by 0.89\% and 1.19\%, while the teacher also yields comparable performance with DML, keeping consistent with our analysis in Sec.\ref{threshold_bound}.

\begin{table*}[t]
\centering
\setlength{\belowcaptionskip}{0.1cm}
\setlength{\abovecaptionskip}{-0.02cm}
\caption{Accuracy (\%) comparison with 3 networks on CIFAR-100. WRN-16-2 serves as either teacher (\textbf{T}) or student (\textbf{S}) for DML and KDCL, while is treated as \textcolor[RGB]{0,255,0}{S} for \textbf{1T2S} and \textcolor[RGB]{0,0,255}{T} for \textbf{2T1S}.   The best results are reported with \textbf{boldface}.}
\tiny
\begin{tabular}{c|c|c|c||c||c}
\toprule
Backbone                  &Vanilla    &DML \cite{zhang2018deep}  &KDCL \cite{Guo_2020_CVPR}   &\tabincell{c}{\quad\textbf{SwitOKD}$\quad$  \\ \textbf{(1T2S)} }   &\tabincell{c}{\quad\textbf{SwitOKD}$\quad$  \\ \textbf{(2T1S)} }        \\

\hline\hline
\rowcolor{mygreen}
MobileNet                  &58.65(\textbf{S})     &63.75(\textbf{S})   &62.13(\textbf{S})   &\textbf{64.62}(\textbf{S})  &\textbf{65.03}(\textbf{S})     \\

WRN-16-2                 &\quad73.37(\textbf{\textcolor[RGB]{0,255,0}{S}/\textcolor[RGB]{0,0,255}{T}})$\quad$     &\quad74.30(\textbf{\textcolor[RGB]{0,255,0}{S}/\textcolor[RGB]{0,0,255}{T}})$\quad$   &\quad73.94(\textbf{\textcolor[RGB]{0,255,0}{S}/\textcolor[RGB]{0,0,255}{T}})$\quad$   &\cellcolor{mygreen}\textbf{75.02}(\textbf{\textcolor[RGB]{0,255,0}{S}})  &71.73(\textbf{\textcolor[RGB]{0,0,255}{T}})   \\

\quad WRN-16-10$\quad$               &79.45(\textbf{T})     &77.82(\textbf{T})   &\textbf{80.71}(\textbf{T})   &77.33(\textbf{T})  &77.07(\textbf{T})    \\
\bottomrule
\end{tabular}
\label{multi_nets}
\end{table*}

\begin{figure*}[t]
\centering
\setlength{\abovecaptionskip}{0.15cm}
\setlength{\belowcaptionskip}{-0.55cm}
\includegraphics[width=0.89\linewidth]{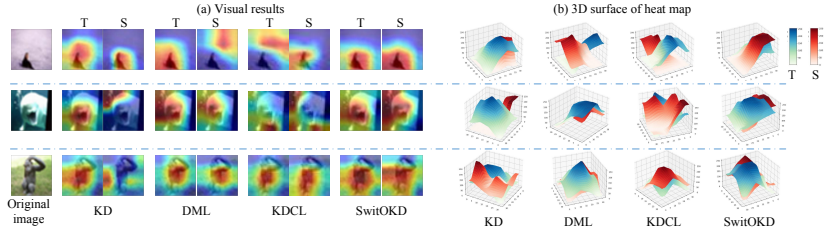}
\caption{Visual analysis of why SwitOKD works on CIFAR-100. 
\textbf{Left:}(a) The visual results by superimposing the heat map onto corresponding original image. \textbf{Right:}(b) 3D surface of heat maps for teacher and student (the more the peak overlaps, the better the student mimics teacher), where $x$ and $y$ axis denote the width and height of an image, while $z$ axis represents the gray value of the heat map.   
T: ResNet-34 (teacher); S: ResNet-18 (student).}
\label{visual_comparison}
\end{figure*}

\subsection{How about SwitOKD from Visualization Perspective?}
To shed more light on why SwitOKD works in Sec.\ref{why_work}, we further perform a visual analysis with Grad-cam \cite{selvaraju2017grad} visualization of image classification via a heat map (red/blue region corresponds to high/low value) that localizes the class-discriminative regions, to confirm that our adaptive switching strategy enables student to mimic teacher well, and thus improves the classification accuracy. 
Fig.\ref{visual_comparison}(a) illustrates the visual results by superimposing the heat map onto corresponding original image, to indicate whether the object regions of the image is focused (the red area denotes more focus); Fig.\ref{visual_comparison}(b) shows 3D surface of the heat map to reflect the overlap range of heat maps for teacher and student (the more the peak overlaps, the better the student mimics teacher). 
Combining Fig.\ref{visual_comparison}(a) and (b), it suggests that SwitOKD focuses on the regions of student, which not only keep consistent with that of the teacher --- mimic the teacher well (KL loss), but  correctly cover the object regions --- yield high precision (CE loss), in line with our analysis (Sec.\ref{introduction}): keep the gradient for KL loss divergent from that for CE loss. \emph{The above further confirms the adversarial impact of large distillation gap --- the emergency of escaping online KD process} (Sec.\ref{tradition} and \ref{learning_m}).

\begin{wrapfigure}{r}{5.25cm}
\centering
\setlength{\belowcaptionskip}{-0.55cm}
\setlength{\abovecaptionskip}{0.2cm}
\includegraphics[width=0.35\textwidth]{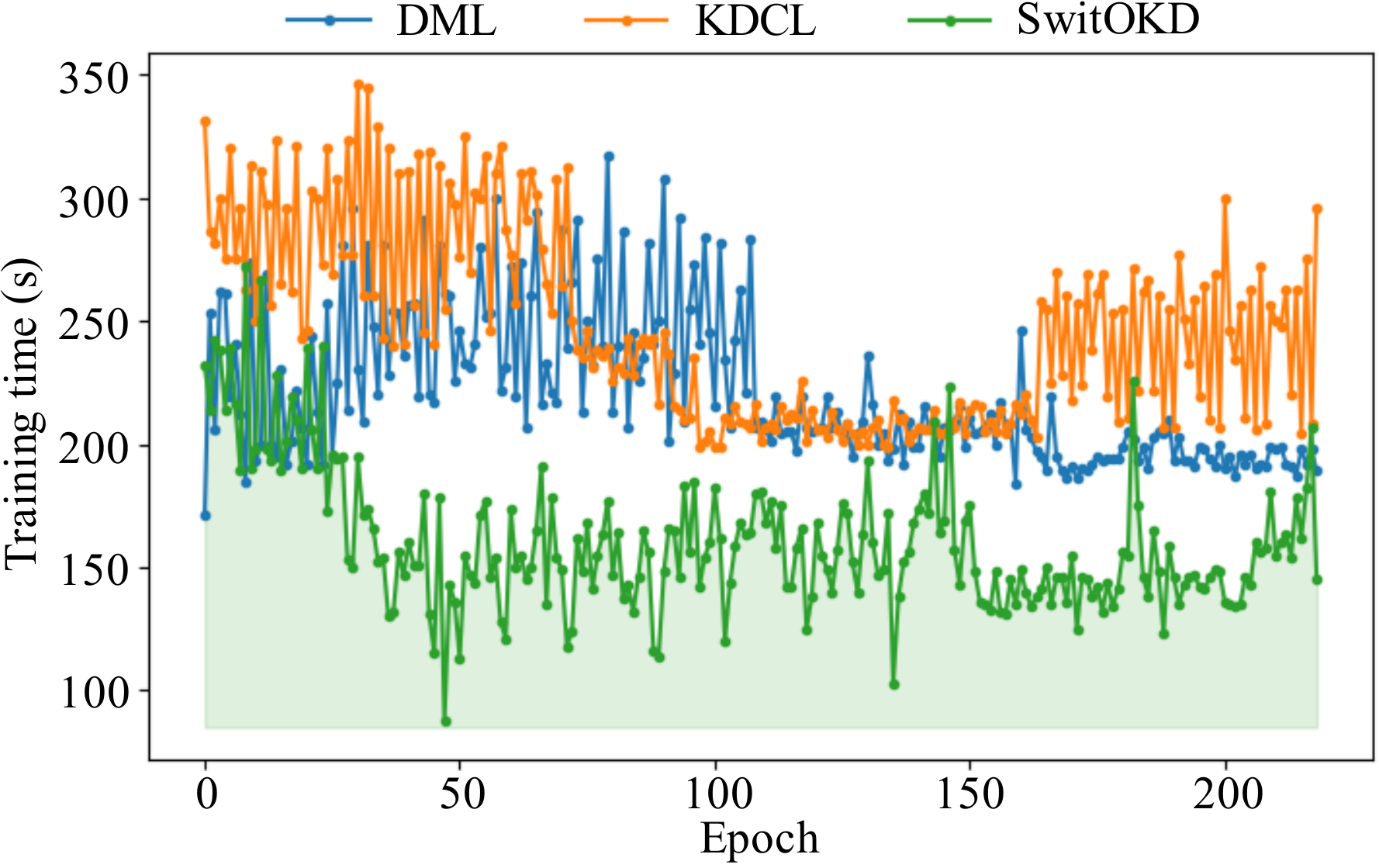}
\caption{Efficiency analysis with MobileNetV2 (student) and ResNet-18 (teacher) on Tiny-ImageNet. }
\label{efficiency}
\end{wrapfigure}

\subsection{What Improves the Training Efficiency of SwitOKD?}
Interestingly, SwitOKD has considerably raised the training efficiency of online distillation process beyond \cite{zhang2018deep,Guo_2020_CVPR} since the training of teacher is \emph{paused} (merely involve inference process) under expert mode (Sec.\ref{expert_m}). We perform efficiency analysis on a single GPU (GTX 1080 Ti), where SwitOKD is compared with other online distillation methods, \emph{e.g.}, DML \cite{zhang2018deep} and KDCL \cite{Guo_2020_CVPR}.
Fig.\ref{efficiency} shows that the time per iteration for SwitOKD (\textcolor[RGB]{44,160,44}{green} line) varies greatly, owing to adaptive switching operation. Notably, the total training time is significantly reduced by 27.3\% ($9.77$h \emph{vs} $13.43$h) compared to DML (\textcolor[RGB]{30,118,180}{blue} line), while 34.8\%  ($9.77$h \emph{vs} $14.99$h) compared to 
KDCL (\textcolor[RGB]{255,127,14}{orange} line).

\subsection{Extension to Multiple Networks}
\label{multi_nets_result}
To show our approach's extendibility for training multiple networks, we conduct the experiments based on three networks with large distillation gap, see Table \ref{multi_nets}. 
As can be seen, the students for \textbf{1T2S} and \textbf{2T1S} achieve significant accuracy gains (5.97\%, 1.65\%,  and 6.38\%) over vanilla and outperform other online distillation approaches (\emph{i.e.}, DML \cite{zhang2018deep} and KDCL \cite{Guo_2020_CVPR}) with significant margins, while our teachers (WRN-16-10) are basically on a par with DML, consistent with the tendency of performance gain for SwitOKD in Table \ref{result_cifar}. By contrast, KDCL receives the best teacher (WRN-16-10), but a poor student (MobileNet), in that it pays more attention to teacher instead of student. Notably, \textbf{1T2S} achieves a better teacher (77.33\% \emph{vs} 77.07\% for WRN-16-10) than \textbf{2T1S}; the reason is that the teacher  for \textbf{1T2S} will be completely suspended when the switching conditions between teacher and both students are triggered (Sec.\ref{multi_n}).